\documentclass[letterpaper, 10 pt, conference]{ieeeconf}
\IEEEoverridecommandlockouts
\usepackage{amsmath, amsfonts, amssymb}
\usepackage{multicol}
\usepackage{color}
\usepackage{graphicx}
\usepackage{subcaption}
\usepackage{dblfloatfix}
\usepackage[hyphens]{url}
\usepackage[bookmarks=true,hidelinks]{hyperref}
\hypersetup{
    pdftitle={OWLS: Open-world Language-conditioned Speed Fields for Zero-Shot Rover Navigation},
    pdfauthor={Chahyon Ku, Erica Tevere, Georgios Georgakis, Rudra Mukherjee, Bernadette Bucher},
    pdfsubject={Robotics},
    pdfkeywords={Robotics, Navigation, Traversability, Vision-language models, Planetary rovers}
}
\usepackage{booktabs}
\usepackage{cite}

\expandafter\ifx\csname iflatexml\endcsname\relax \expandafter\newif\csname iflatexml\endcsname \fi

\begin{document}
\bstctlcite{IEEEexample:BSTcontrol}

\title{\LARGE \bf OWLS: Open-world Language-conditioned Speed Fields \\ for Zero-Shot Rover Navigation}

\iflatexml
\author{\IEEEauthorblockN{Chahyon Ku}\IEEEauthorblockA{Department of Robotics, University of Michigan\\ {\tt chahyon@umich.edu}}
\and \IEEEauthorblockN{Erica Tevere}\IEEEauthorblockA{Jet Propulsion Laboratory, California Institute of Technology}
\and \IEEEauthorblockN{Georgios Georgakis}\IEEEauthorblockA{Jet Propulsion Laboratory, California Institute of Technology}
\and \IEEEauthorblockN{Rudra Mukherjee}\IEEEauthorblockA{Jet Propulsion Laboratory, California Institute of Technology}
\and \IEEEauthorblockN{Bernadette Bucher}\IEEEauthorblockA{Department of Robotics, University of Michigan\\ {\tt bucherb@umich.edu}}}
\else
\author{Chahyon Ku$^{1}$, Erica Tevere$^{2}$, Georgios Georgakis$^{2}$, Rudra Mukherjee$^{2}$, and Bernadette Bucher$^{1}$\\[3pt]
{\sublargesize Project Page: {\fontfamily{zi4}\selectfont\urlstyle{same}\textcolor[rgb]{0.212,0.490,0.741}{\url{https://chahyon-ku.github.io/owls/}}}}%
\thanks{A portion of this research was carried out at the Jet Propulsion Laboratory, California Institute of Technology, under a contract with the National Aeronautics and Space Administration (80NM0018D0004).}%
\thanks{$^{1}$C. Ku and B. Bucher are with the Department of Robotics, University of Michigan, Ann Arbor, MI 48109, USA {\tt\small \{chahyon, bucherb\}@umich.edu}}%
\thanks{$^{2}$E. Tevere, G. Georgakis, and R. Mukherjee are with the Jet Propulsion Laboratory, California Institute of Technology, Pasadena, CA 91109, USA}%
}
\fi

\maketitle
\iflatexml
\begin{center}Project Page: \url{https://chahyon-ku.github.io/owls/}\end{center}
\fi
\begin{abstract}
Semantics can provide a proxy for physical traversability. Terrain that looks like loose sand or a rock field is likely to have similar physical properties. Open-vocabulary vision-language models (VLMs) enable this physical reasoning over terrain outside any fixed label set.
We present Open-world Language-conditioned Speed Fields (OWLS), which plans directly over the dense feature map of a VLM.
Prompted with a description of traversable terrain, the VLM scores every pixel of a single camera image; OWLS projects these scores onto the ground, adapts them to the platform with two morphological operations, and uses the result as the speed field of a minimum-time planner, so terrain appearance and obstacle clearance shape one trajectory.
OWLS needs no task-specific training, no terrain annotation, no persistent map, and no exteroceptive sensor beyond the camera.
We evaluate it on a physical rover indoors and off-road, on simulated Mars and Moon rovers over 200 procedurally generated terrains, and on real planetary imagery.
\end{abstract}

\section{Introduction}
\label{sec:introduction}
Semantics can provide a proxy for physical traversability. Terrain that looks like loose sand or a rock field is likely to have similar physical properties. Whether a robot can cross a stretch of terrain depends on information such as the size of obstacles relative to its footprint and clearance, load bearing, and slip, which a camera cannot measure directly. 
Planetary rovers must reason about these physical properties under severe conditions. Rovers drive without GPS, and communication delays force long stretches of onboard autonomy~\cite{enav}.

Existing systems model physical traversability in several ways.
Hand-designed geometric heuristics, such as step-height or slope limits, require substantial per-platform engineering~\cite{step}.
Learned terrain models~\cite{bevnet,terrainnet,seeing_through_grass} predict a fixed set of classes from annotated data, and each class still needs a hand-assigned planning cost, so neither the classes nor the costs cover terrain the dataset did not anticipate.
Self-supervised estimators learn traversability from the robot's own driving~\cite{wvn,velociraptor}, which presumes experience on the target terrain.
Open-vocabulary vision-language models (VLMs) remove the fixed label set, but navigation systems built on them take free space from depth or LiDAR and use the VLM to select a goal, rank candidate paths, or annotate that geometric map~\cite{vlmaps,vlfm,vlmnav,behav,opennav}.
None of these existing approaches plan over the VLM's dense output itself, although that output already scores every pixel against a description of drivable terrain.
\begin{figure}[!t]
\centering
\includegraphics[width=\columnwidth]{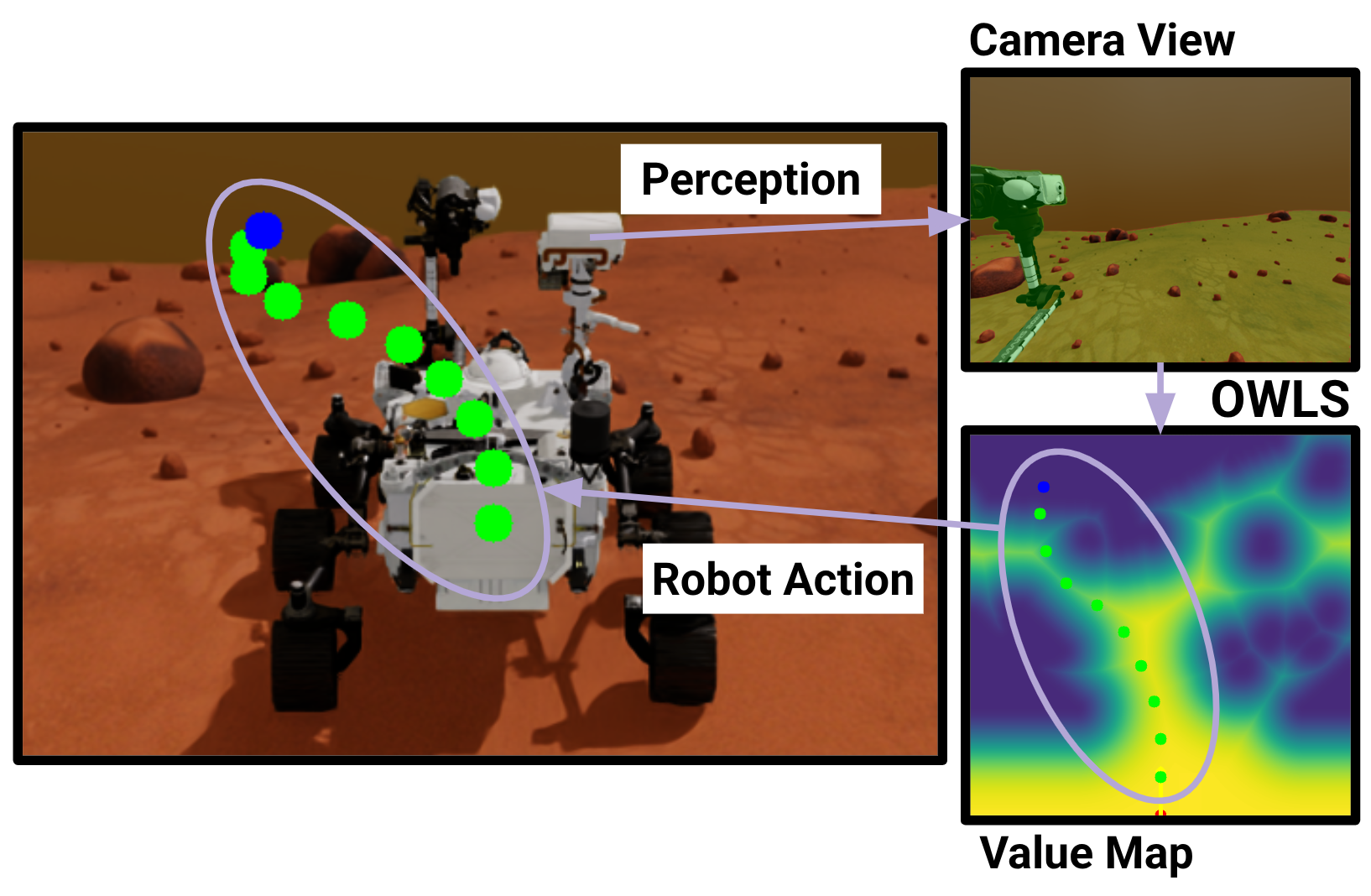}
\caption{OWLS plans directly over the dense feature map of a VLM: the VLM scores a single camera image against a description of traversable terrain (green), and the scores become the speed field of a minimum-time planner whose waypoints (green) the rover follows to the goal (blue).}
\label{fig:teaser}
\end{figure}

We present Open-world Language-conditioned Speed Fields (OWLS), a local navigation method that plans directly over the dense feature map of an open-vocabulary VLM (Fig.~\ref{fig:teaser}).
Prompted with a description of traversable terrain, the VLM scores each pixel of the camera image, and OWLS turns this score map into a trajectory in three steps.
Ground projection places each score at a metric position in the robot frame.
Two morphological operations then encode the platform, since a semantic score does not say whether a particular robot can drive over a rock or how far it must keep from one: a dilation removes low-scoring regions small enough to drive over, and an erosion surrounds the rest with a clearance buffer scaled to the vehicle.
Finally, rather than thresholding the result into occupancy, OWLS uses it as the speed at which the robot moves through each cell and plans the minimum-time trajectory to the goal, which a waypoint controller tracks.
Obstacle clearance and the VLM's judgment of the terrain thus enter a single optimization.
OWLS uses the VLM without fine-tuning and needs no terrain annotation, and adapting to a new rover needs at most two radii set from the rover's dimensions.
It keeps no persistent map, replanning from each new image with a single camera and wheel odometry.

\noindent
\textbf{Contributions.}
(i) We show how to \textbf{plan directly over the dense feature map of an open-vocabulary VLM}, using ground-projected VLM scores as the speed field of a minimum-time planner, which yields local navigation over previously unseen terrain from a single camera with no task-specific training and no persistent map.
(ii) We introduce \textbf{an interpretable embodiment adaptation} of the VLM map by two morphological operations with radii set from rover geometry, which carries the same VLM and prompt across three rovers of distinct size and camera placement without retraining.
(iii) We \textbf{evaluate from terrestrial to planetary settings}: on a physical rover, on a simulated rover on 200 procedurally generated Mars and Moon terrains with prompt and morphology ablations, and qualitatively on real planetary imagery.

\section{Related Work}
\noindent
\textbf{Traversability from Appearance.}
Traversability is a physical property, and existing methods differ in how they reach it.
Geometric methods measure it from range data: Perseverance checks stereo-derived terrain against the vehicle's clearance and tilt limits~\cite{enav}, and STEP~\cite{step} evaluates risk from LiDAR geometry, both with substantial per-platform engineering.
Semantic methods use appearance as a proxy for the same properties and learn the mapping from appearance to cost.
The common instance trains terrain classifiers over a fixed label set on annotated data~\cite{bevnet, terrainnet, seeing_through_grass} and assigns each class a planning cost; annotation is expensive, per-class costs are hard to design, and the closed class set is often ambiguous or application-specific.
Other methods ground the proxy in the robot's own experience, through vibration-based ground classification~\cite{rnn_proprioceptive}, wheel-slip distributions with uncertainty for risk-aware planning~\cite{evora}, traversability learned from the robot's driving over pretrained visual foundation features~\cite{wvn, velociraptor}, or a segmentation foundation model distilled with egocentric video~\cite{semantic_traversability_egocentric}.
These remove manual labels but still train a platform-specific estimator.
OWLS takes the appearance proxy from a pretrained VLM's response to a text description, with no training, and moves the platform dependence out of the estimator into two morphological radii set from rover geometry.

\noindent
\textbf{Mapless Navigation and Planning over Speed Fields.}
Learned policies act on RGB images without a globally consistent map, trained by imitation~\cite{driving_imiatation_learning} or offline reinforcement learning~\cite{offline_rl_visual_navigation}; GNM~\cite{gnm}, ViNT~\cite{vint}, and NoMaD~\cite{nomad} train on large cross-platform corpora of human-collected navigation data for image-goal-conditioned navigation and unconditioned exploration.
They model traversability only implicitly, are bounded by the diversity of their data, and offer no way to state which terrain counts as drivable.
Classical visual servoing also closes the loop on each new image without a map~\cite{references:tutorialvisualservoing, mobile_vs_1}, but on hand-engineered geometric features~\cite{harris, visp}.
OWLS instead keeps an explicit, interpretable planner.
Eikonal planners compute minimum-time paths under a speed field with fast marching~\cite{fmm} or the Fast Iterative Method~\cite{fim}, and mobile robot planners derive that speed from occupancy or distance to obstacles~\cite{fm2}.
OWLS derives it from a VLM feature map, so terrain appearance, not only geometry, sets how fast the robot can move through each cell.

\noindent
\textbf{VLMs for Navigation.}
Open-vocabulary VLMs supply the appearance proxy without a fixed label set, but navigation systems built on them take free space from other sensors.
VLMaps~\cite{vlmaps} fuses open-vocabulary features into a metric map built from depth and uses language to pick goals; it also derives per-robot obstacle maps by listing obstacle categories in text, a closed list per embodiment.
VLFM~\cite{vlfm} builds a language-grounded value map from a pretrained VLM, as OWLS does, but uses it to rank the frontiers of an occupancy map built from depth.
VLMNav~\cite{vlmnav} takes traversable regions from perfect simulated depth, samples candidate paths, and asks a VLM to select one.
BehAV~\cite{behav} and Open-Nav~\cite{opennav} plan on occupancy grids from LiDAR and use VLM masks to set a goal or encode operator preferences (e.g.\ ``avoid grass'' although grass is traversable), BehAV requiring the operator to enumerate object types and their desirability per scenario.
VAMOS~\cite{vamos} trains a planner on diverse data to propose paths in image space and re-ranks them with a per-embodiment affordance model learned in simulation.
In each system the VLM selects, annotates, or proposes, while what the robot can traverse comes from depth, LiDAR, or a learned model.
OWLS plans over the VLM's dense output itself: from a single calibrated camera and with no training, the ground-projected feature map, adapted by two radii, is the speed field of the planner.

\section{Method}
\label{sec:method}

\begin{figure*}[t!]
\centering
\includegraphics[width=0.68\textwidth]{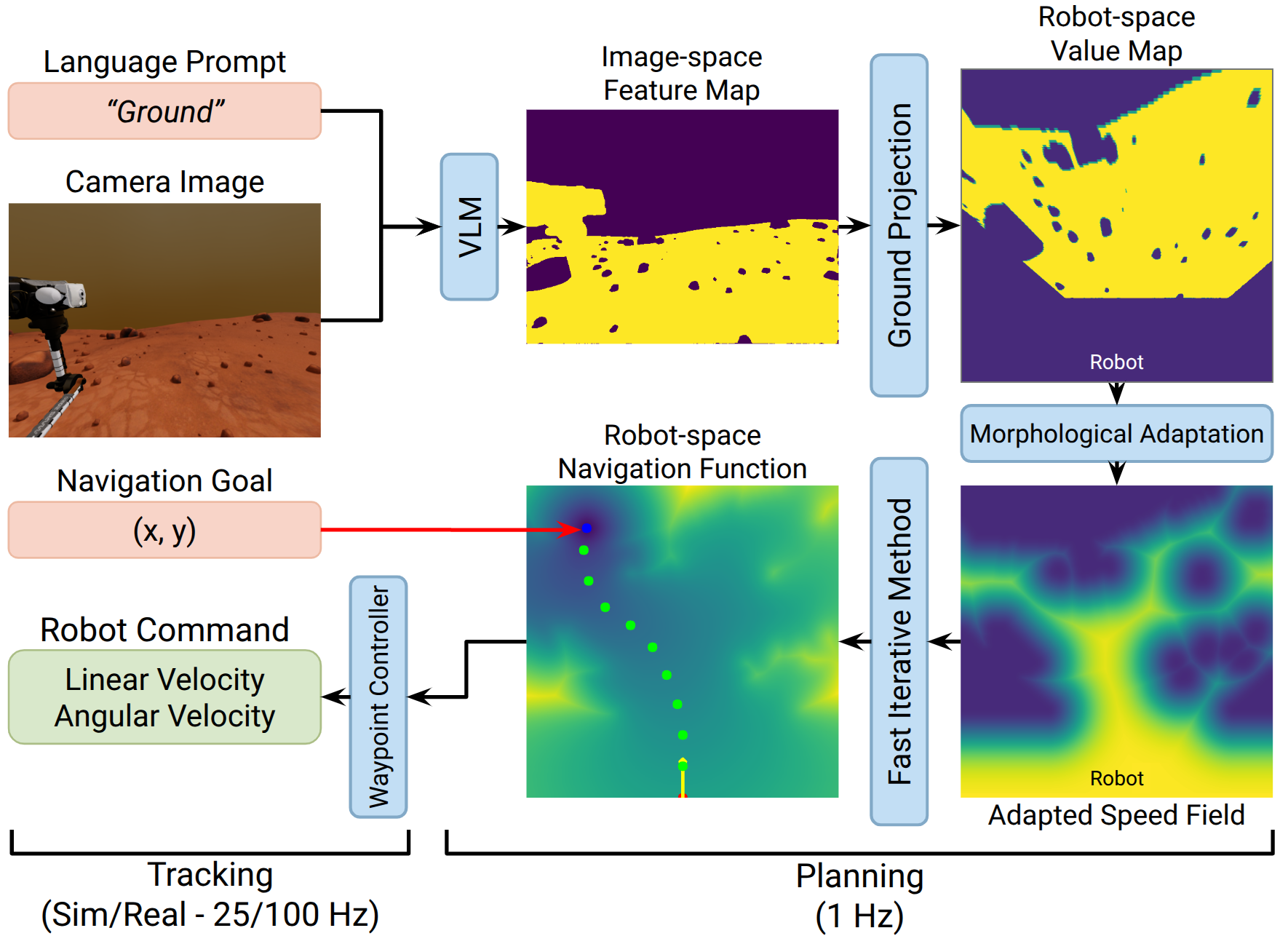}
\caption{OWLS plans directly over the dense feature map of a VLM (Sec.~\ref{sec:planning}). The VLM scores each pixel against the prompt; the scores are ground-projected into the robot frame, adapted to the platform by dilation and erosion, and used as the speed field of a minimum-time planner, whose trajectory a waypoint controller tracks (Sec.~\ref{sec:tracking}).}
\label{fig:system_diagram}
\end{figure*}

OWLS plans a local trajectory directly over the dense feature map of an open-vocabulary VLM and tracks it with a waypoint controller, replanning from each new image without a persistent map (Fig.~\ref{fig:system_diagram}).
Planning (Sec.~\ref{sec:planning}) runs at the perception rate: the VLM scores each pixel against a text prompt, ground projection lifts the scores into a robot-frame value map $V$, two morphological operations adapt $V$ to the platform, and the adapted map sets the speed field of a minimum-time planner.
Tracking (Sec.~\ref{sec:tracking}) runs at the control rate and steers the robot along the planned waypoints.

\subsection{Planning over VLM Features}
\label{sec:planning}

\noindent
\textbf{Language-conditioned Traversability.}
We prompt an open-vocabulary VLM with a description of traversable terrain and read out its dense, pixel-aligned response $H \in [0,1]^{h\times w}$, which we call its feature map.
For open-vocabulary segmenters such as Grounded SAM~2 \cite{grounding_dino, grounded_sam, sam2} and SAM~3 \cite{sam3}, $H$ is the pixel-wise mask probability before thresholding.
We keep the continuous response rather than a binary mask, so the planner can weigh how strongly each pixel matches the prompt.
Every simulated experiment uses the single prompt ``\texttt{ground}'', held fixed across platforms, VLMs, environments, and episodes, so that per-platform adaptation is carried by the morphological radii below; the hardware trials of Sec.~\ref{sec:real_rover_experiments} use ``\texttt{drivable ground}''.

\noindent
\textbf{Ground Projection.}
Clearance and travel time are metric quantities in the robot frame, so we lift $H$ onto the ground to form a top-down value map $V$.
We assume flat ground, so that terrain within the local window lies on the plane $z=0$ of the robot frame, and, with intrinsics $\mathbf{K}$ and calibrated robot-to-camera extrinsics, a ground point projects through the planar homography $\mathbf{M}$ obtained by dropping the third column of the extrinsic matrix.
Each cell of a regular grid $\Omega\subset\mathbb{R}^2$ takes the value of $H$ at its projection through $\mathbf{M}$, or $0$ if it projects outside the image or above the horizon, giving $V:\Omega\to[0,1]$.

\noindent
\textbf{Morphological Adaptation.}
The VLM scores how well terrain matches the prompt, not whether a given robot can traverse it, so we adapt $V$ to the platform's size and capability with two grayscale morphological operations.
A \textit{dilation} with a disk of radius $\rho_t$ erases low-traversability regions, including VLM artifacts, smaller than the largest obstacle the platform can drive over.
An \textit{erosion} with a Gaussian-tapered disk of radius $\rho_r$ then grows the remaining ones into a clearance buffer whose penalty decays with distance, giving $V'$.
Neither operation thresholds the map, so traversability stays continuous.
Both radii follow from rover geometry (detailed later in Sec.~\ref{sec:implementation}): $\rho_r$ is set per platform from its track width and $\rho_t$ is shared across platforms.

\noindent
\textbf{Speed Field and Minimum-time Trajectory.}
Because $V'$ is continuous, we plan over it directly instead of thresholding it into occupancy.
We take $c = s\,V' + \epsilon$ as the speed at which the robot moves through each cell, generalizing the binary navigation function, in which free cells have uniform speed and obstacles have zero speed.
The floor $\epsilon > 0$ keeps every cell passable, so the robot plans through low-traversability terrain only where going around takes longer, and unobserved cells ($V=0$) move at the same floor as confident obstacles.
The navigation function $N$ is the travel time from each cell to the goal under $c$, the solution of the eikonal equation $\lVert\nabla N\rVert = 1/c$, computed with the Fast Iterative Method \cite{fim}.
It is seeded at the goal cell when the goal lies in $\Omega$ and otherwise at the boundary cell along the goal bearing, clipped to a forward cone; the seed is recomputed each cycle, so the effective goal recedes as new terrain enters view.
The trajectory $\mathbf{f} = ((0,0),(x_1,y_1),\ldots)$, in the robot frame at image capture, descends the navigation function $N$ from the robot: from each point we take the step within a fixed radius that most reduces the time to goal.
It is the minimum-time path under the VLM-derived speed field, so the clearance buffers shape it around obstacles and the VLM scores steer it across terrain, in one optimization.
We use a minimal controller (Eq.~\eqref{eq:control}) to focus on the contribution of our speed field, which suffices for low-speed tracking in our simulated and real-rover experiments, but we can swap it for more advanced control algorithms in deployment scenarios requiring more complex dynamics.

\subsection{Trajectory Tracking}
\label{sec:tracking}

A new trajectory arrives with each image; between images, wheel odometry transforms it into the robot's current frame.
The controller steers toward the first waypoint $\mathbf{p}^{*}=(x_i,y_i)$ farther than a lookahead distance $d_{\min}$, at bearing $e_{\mathrm{ang}}=\operatorname{arctan2}(y_i,x_i)$.
With an alignment threshold $\alpha$ and range $r_g$ to the commanded goal, it switches between turning in place and driving straight,
\begin{align}
  w^{*} &= w_{\max}\,\operatorname{sign}(e_{\mathrm{ang}})\,\mathbf{1}\{|e_{\mathrm{ang}}| \ge \alpha\}\,\mathbf{1}\{r_g \ge d_{\mathrm{goal}}\}, \\
  v^{*} &= v_{\max}\,\mathbf{1}\{|e_{\mathrm{ang}}| < \alpha\}\,\mathbf{1}\{r_g \ge d_{\mathrm{goal}}\},
  \label{eq:control}
\end{align}
and halts within $d_{\mathrm{goal}}$ of the goal.
On a mobile RTX 4090, a planning cycle with GSAM2 takes $692$\,ms, $490$\,ms of it in the VLM and $199$\,ms in the morphology, consistent with the $1$\,Hz rate; waypoint and action generation are sub-millisecond.

\subsection{Implementation Details}
\label{sec:implementation}

\noindent
\textbf{Blind and Self-occluded Cells.}
Two kinds of cells show no terrain without being obstacles: the blind strip between the robot and the nearest visible ground, and cells hidden behind fixed vehicle parts in view, such as Perseverance's stowed arm.
Both are set to maximum traversability instead of $V=0$, so they do not read as obstacles.

\begin{figure}[t!]
\centering
\includegraphics[width=1.0\linewidth]{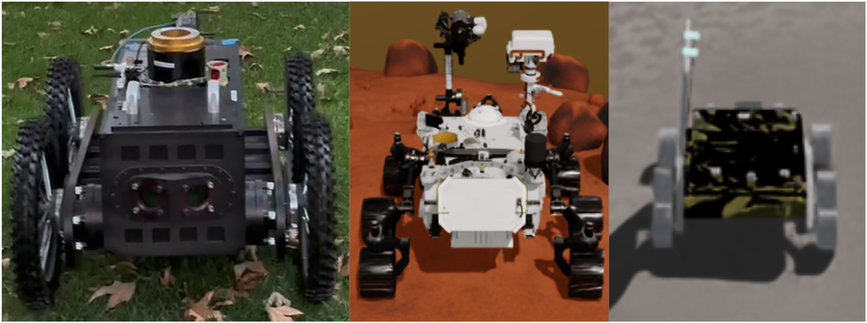}
\caption{Real 4-wheeled rover (left), simulated Perseverance (middle), and simulated Pragyan (right) used in the experiments.}
\label{fig:robots}
\end{figure}

\begin{figure*}[t!]
\centering
\iflatexml
\includegraphics[width=\linewidth]{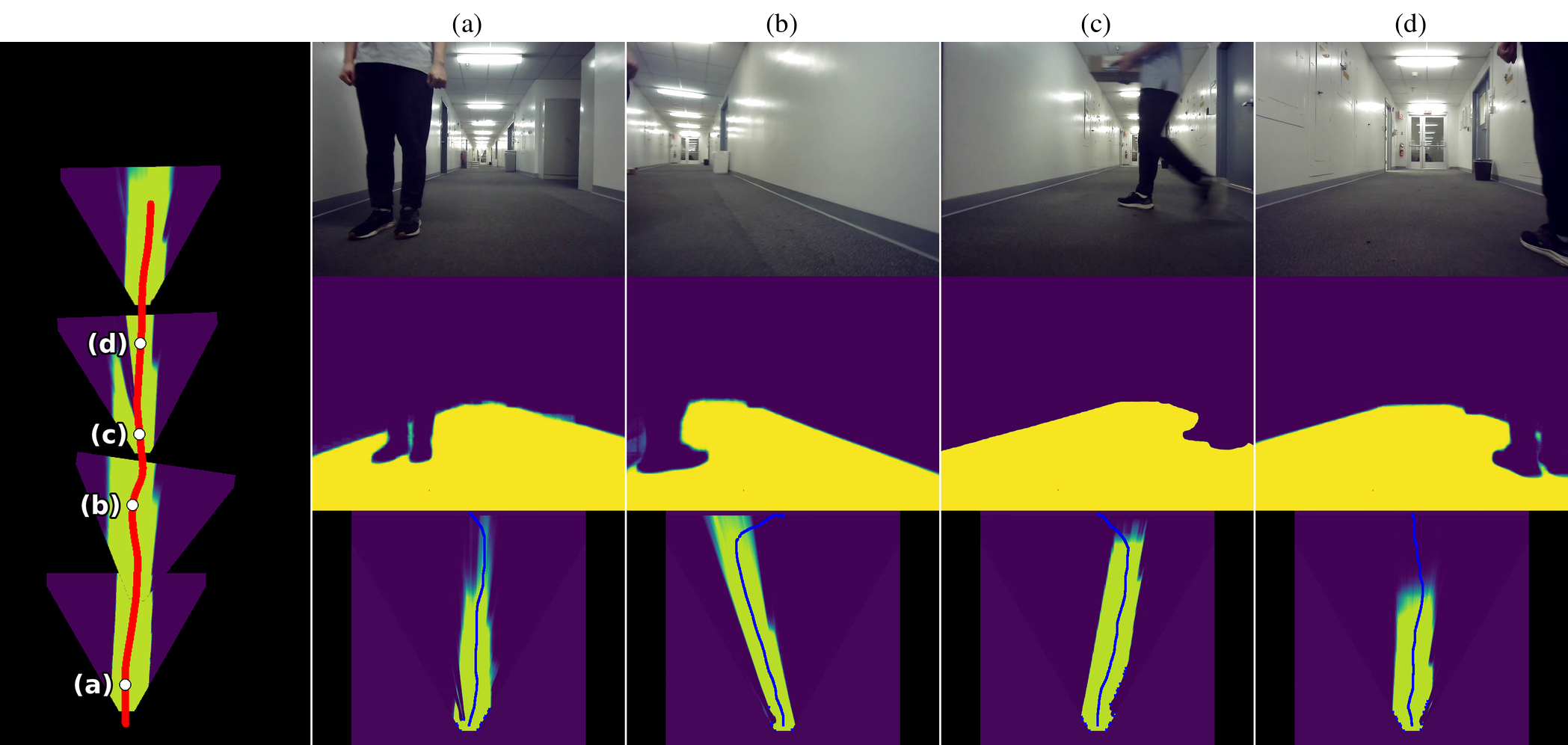}
\else
\begin{minipage}{0.90\textwidth}
\centering\footnotesize
\makebox[0.1980\linewidth]{}\makebox[0.2005\linewidth]{(a)}\makebox[0.2005\linewidth]{(b)}\makebox[0.2005\linewidth]{(c)}\makebox[0.2005\linewidth]{(d)}\\[1pt]
\includegraphics[width=0.95\linewidth]{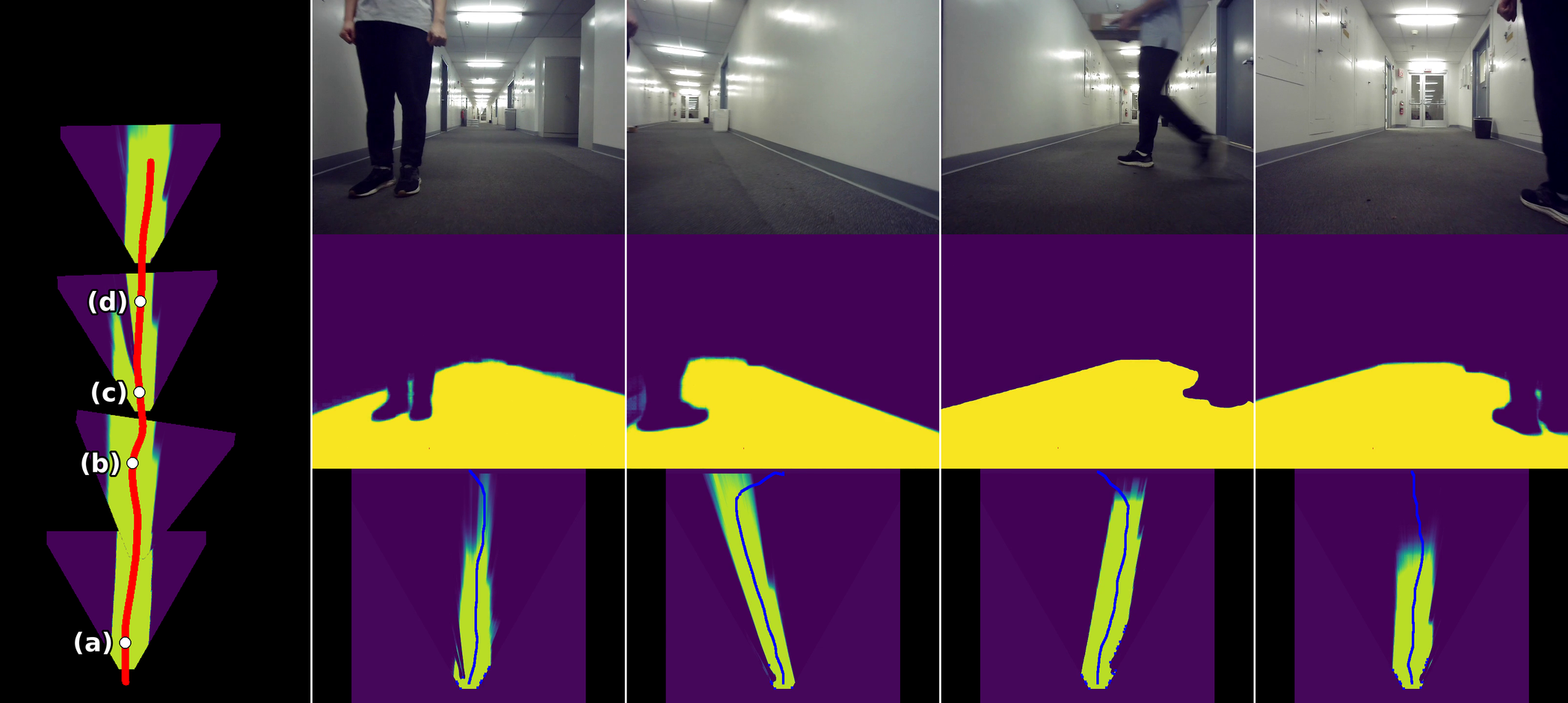}
\end{minipage}
\fi
\caption{
Physical rover navigating an indoor hallway while a person walks across its path (Sec.~\ref{sec:real_rover_experiments}). Left: executed trajectory (red) over the top-down traversability of successive perception cycles. (a--d) The four timesteps, each with the camera view (top), the VLM's image-space heatmap (middle), and the top-down speed field with the planned trajectory (blue, bottom). The heatmap excludes each person as they cross, and since OWLS keeps no map, each cycle replans over the ground left unoccupied at that moment.
}
\label{fig:real_dynamic}
\end{figure*}

\noindent
\textbf{Parameters.}
The grid $\Omega$ is $640\times640$ cells at $0.0125$\,m, spanning $8$\,m ahead of the robot and $\pm4$\,m laterally.
The speed field uses $\epsilon=0.1$ and a VLM-dependent gain, $s=8$ for Grounded SAM~2 and $s=16$ for SAM~3, %
which corrects for the different ranges of the two models' raw mask probabilities $H$.
The morphological radii are $\rho_r = 2.15$\,m for Perseverance and $0.50$\,m for Pragyan and the terrestrial rover, with $\rho_t = 0.19$\,m throughout.
The controller uses a lookahead $d_{\min}$ and stopping radius $d_{\mathrm{goal}}$ both of $0.25$\,m, and $\alpha = 8^{\circ}$ on all platforms, with linear and angular speeds $(v_{\max}, w_{\max})$ of $(0.70,\,2.06)$, $(0.35,\,2.19)$, and $(0.86,\,2.42)$ in m/s and rad/s for Perseverance, Pragyan, and the terrestrial rover.
Planning runs at $1$\,Hz on all platforms, and control at $25$\,Hz in simulation and $100$\,Hz on the physical rover.

\section{Experiments}
\label{sec:experiments}

Lacking access to a real space rover, we characterize monocular planetary navigation in four steps: real-rover trials (Sec.~\ref{sec:real_rover_experiments}) establish feasibility under real compute and sensor constraints; a simulated sweep over 100 terrains per platform (Sec.~\ref{sec:simulated_rover_experiments}) tests transfer to space embodiments and appearance; ablations over the same terrains (Sec.~\ref{sec:sensitivity}) vary the prompt and the morphology, the only inputs set per deployment; and inference on real space-rover imagery (Sec.~\ref{sec:offline_inference}) tests the VLM under flight lighting and sensor noise.

\subsection{Physical Rover Experiments on Earth}
\label{sec:real_rover_experiments}

To test our method under real compute and sensor constraints, we run it on a 4-wheeled rover in two Earth scenarios: an indoor hallway and an off-road field mixing grass, dirt, bushes, trees, and pavement.

\noindent
\textbf{Setup.}
The rover carries a single front-facing RGB pinhole camera at $640 \times 480$, and our method runs on a laptop with a mobile NVIDIA RTX 4090, generating a trajectory from the front-view image and publishing closed-loop wheel velocity commands at the rates of Sec.~\ref{sec:implementation}.
Wheel encoders provide odometry at the control rate, closing the velocity loop and carrying the trajectory forward between perception updates (Sec.~\ref{sec:tracking}).
All trials ran the prompt ``\texttt{drivable ground}''.
Because the planner accepts any dense, pixel-aligned VLM output, we compare three feature maps to plan over: \textbf{Grounded SAM~2} (GSAM2) \cite{grounding_dino, grounded_sam, sam2}, \textbf{DINOv3 patch embeddings} (DINOv3-patch) \cite{dinov3}, and \textbf{DINOv3 language embeddings} (DINOv3-lang) \cite{dinov3, dinov2txt}.
For both DINOv3 variants the feature map is the similarity of patch features to a query embedding: DINOv3-patch queries with the image patch ahead of the rover and takes no text prompt, serving as a language-free control; DINOv3-lang queries with a text embedding, applying the DINOv2-txt \cite{dinov2txt} recipe to DINOv3, as released.

\begin{table}
    \centering
    \begin{tabular}{ccccc}
        \toprule
        Environment & Obstacles & GSAM2 & \shortstack{DINOv3\\patch} & \shortstack{DINOv3\\lang} \\
        \midrule
        Indoor & Static & 100 & 40 & 0 \\
        Indoor & Dynamic & 80 & 20 & 0 \\
        Outdoor & Static & 100 & 40 & 0 \\
        Outdoor & Dynamic & 40 & 0 & 0 \\
        \bottomrule
    \end{tabular}
    \caption{Navigation success rate (\%) of a real 4-wheeled rover in two Earth scenarios with dynamic / static obstacles over 5 randomized trials per configuration (Sec.~\ref{sec:real_rover_experiments}).}
    \label{table:real_success_rates}
\end{table}

\noindent
\textbf{Metric.}
We report navigation success rate over 5 randomized trials per configuration, a trial succeeding if the robot traverses the environment end to end without collision.
This count establishes feasibility rather than fine-grained comparison, so we treat differences of one or two trials as indicative only; it measures how consistently the method follows traversable area up to a local dead-end.

\noindent
\textbf{Is OWLS capable of driving a real mobile robot in various environments?}
With GSAM2, OWLS avoided all static obstacles indoors and outdoors in all 5 trials (Table~\ref{table:real_success_rates}), and dynamic ones such as humans and boxes indoors 80\% of the time (Fig.~\ref{fig:real_dynamic}); in dynamic outdoor scenes only 2 of 5 succeeded.
We attribute these to under-segmentation in unconstrained outdoor lighting plus phase lag from the two-rate architecture.
DINOv3-patch succeeded less often (Table~\ref{table:real_success_rates}): its smoother feature map made it less reactive, so it avoided only obstacles that small steering corrections could clear.
DINOv3-lang traversed nothing: on our trial footage its text query spans roughly a seventh of the patch query's raw similarity range, which per-frame renormalization then amplifies.

\subsection{Simulated Rover Experiments on Mars and Moon}
\label{sec:simulated_rover_experiments}

To verify whether performance on a 4-wheeled Earth rover transfers to space rovers (different embodiment) in space environments (different appearance and obstacles), we evaluate in the Space Robotics Bench (SRB) \cite{space_robotics_bench}.

\noindent
\textbf{Setup.}
SRB provides robot definitions and procedurally generated Mars and Moon terrain over IsaacLab \cite{isaaclab} and the IsaacSim \cite{isaacsim} renderer.
We use the provided Perseverance and Pragyan definitions (Fig.~\ref{fig:robots}) and terrains as is, adding a stationary target $10$\,m away at a randomly drawn bearing.
Only Perseverance needs the self-occlusion mask of Sec.~\ref{sec:implementation}, for its stowed arm.
We run OWLS with two VLMs: \textbf{GSAM2}~\cite{grounding_dino, grounded_sam, sam2}, as in the real-rover study, and the more recent \textbf{SAM~3}~\cite{sam3}.
They share every geometric, morphological, and control parameter; only the speed gain is set per VLM (Sec.~\ref{sec:implementation}), so the comparison isolates the VLM with its calibration.
We do not carry the DINOv3 variants into simulation, since DINOv3-patch proved insufficiently reactive and DINOv3-lang failed outright during the hardware experiments.
All simulated runs use the prompt ``\texttt{ground}'' and the radii of Sec.~\ref{sec:implementation}.

\noindent
\textbf{Baseline.}
We compare against NoMaD \cite{nomad}, a navigation foundation model trained on ``over 100 hours'' of heterogeneous navigation data from varied robots and environments \cite{gnm, sacson}.
Where OWLS plans one trajectory over the VLM feature map, NoMaD samples many unconditioned trajectories; following the same conventions used in VAMOS \cite{vamos}, we take the one whose waypoint is closest to the goal.
Neither method is fine-tuned in domain, as both are meant to run out of the box.
NoMaD has no interface for a metric goal or platform calibration, so its network runs unconditioned, the goal entering only through the selection above, where OWLS receives the goal, calibration, odometry, and its two morphological radii.
We read it as a reference point for a learned navigation prior rather than a matched baseline.

\noindent
\textbf{Metric.}
We report \textit{navigation success rate}: success is holding within $0.25$\,m of the target for one second inside a $60$\,s episode, and a no-progress guard ends an episode early if the rover gains less than $0.1$\,m on the goal over a $20$\,s window, both applied identically to all methods.
The criterion scores goal-reaching alone and carries no collision term, so contact with a rock is penalized only indirectly, on the episodes where it stalls the rover into the no-progress guard.
Mean completion time over successes conditions each method on a different terrain subset, so for efficiency we use success-weighted time efficiency, $\mathrm{STE} = \frac{1}{N}\sum_i S_i\, t_{\mathrm{opt}}/\max(t_i, t_{\mathrm{opt}})$, with $S_i$ success on terrain $i$, $t_i$ its time, and $t_{\mathrm{opt}} = 10\,\text{m}/v_{\max}$ ($14.3$ and $28.6$\,s for Perseverance and Pragyan).
Averaging over all $N$ terrains keeps STE comparable across methods, in the spirit of success weighted by path length (SPL) \cite{spl}.
SRB fixes its procedural terrain per process, so to sample the obstacle distribution we launch one process per terrain, 100 per platform with one episode each, with the start pose and goal direction fixed across terrains and conditions so that terrain is the only quantity that varies.
Every condition runs on the same 100 terrains.

\begin{table}[t]
    \centering
    \setlength{\tabcolsep}{4pt}
    \begin{tabular}{lcc}
        \toprule
        Method & Success Rate (\%) & STE \\
        \midrule
        \multicolumn{3}{l}{\textit{Perseverance (Mars)}} \\
        NoMaD \cite{nomad}  & 23.0 & 0.119 \\
        OWLS + GSAM2 (ours)  & 58.0 & 0.319 \\
        OWLS + SAM~3 (ours)  & 60.0 & 0.319 \\
        \midrule
        \multicolumn{3}{l}{\textit{Pragyan (Moon)}} \\
        NoMaD \cite{nomad}  & 35.0 & 0.217 \\
        OWLS + GSAM2 (ours)  & 75.0 & 0.561 \\
        OWLS + SAM~3 (ours)  & 77.0 & 0.597 \\
        \bottomrule
    \end{tabular}
    \caption{Simulated space experiments (Sec.~\ref{sec:simulated_rover_experiments}) over the same 100 terrains per platform. On terrains both solve, NoMaD takes $30.4$ vs.\ $27.3$\,s against OWLS with GSAM2 on Perseverance and $46.4$ vs.\ $38.6$\,s on Pragyan.}
    \label{table:simulated_space_experiments}
\end{table}

\begin{figure}[tb!]
\centering
\iflatexml
\includegraphics[width=\linewidth]{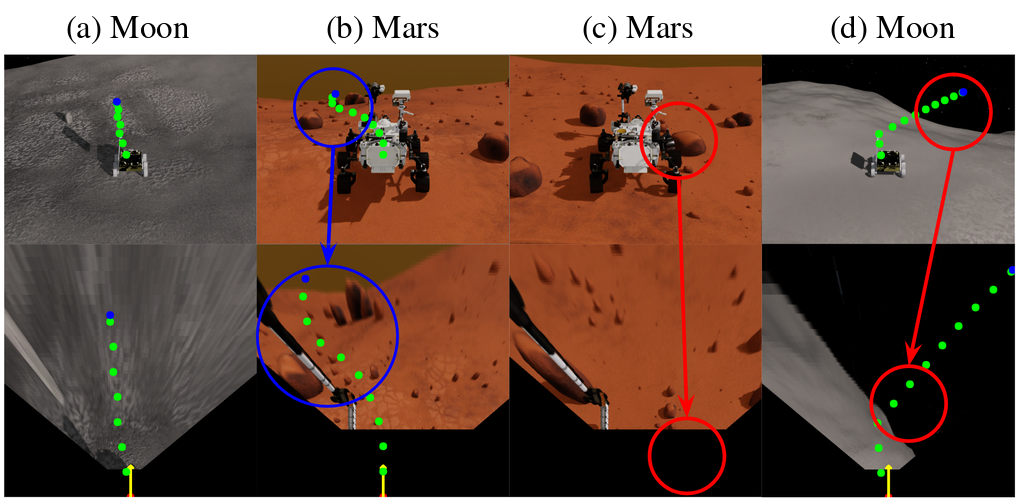}
\else
\begin{minipage}{\linewidth}
\centering\footnotesize
\makebox[0.25\linewidth]{(a) Moon}\makebox[0.25\linewidth]{(b) Mars}\makebox[0.25\linewidth]{(c) Mars}\makebox[0.25\linewidth]{(d) Moon}\\[1pt]
\includegraphics[width=\linewidth]{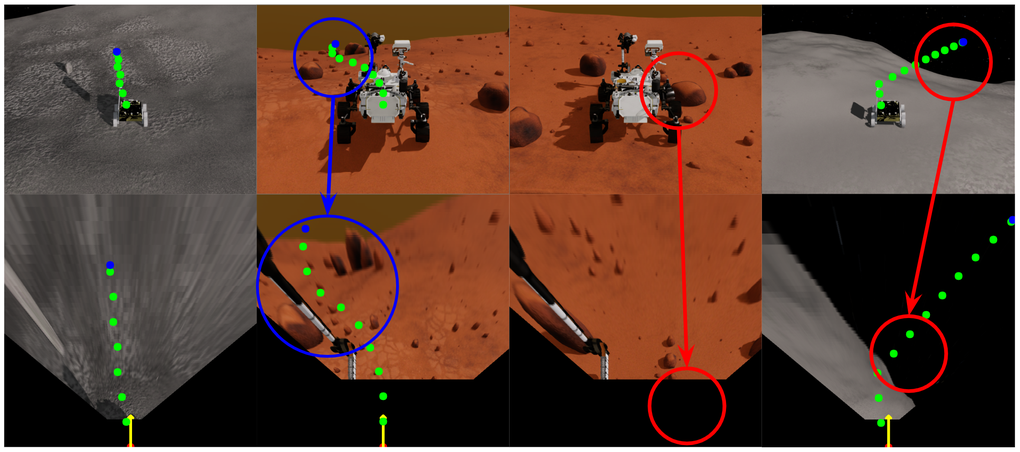}
\end{minipage}
\fi
\caption{
Simulation results: third-person view (top, visualization only) and top-down RGB (bottom, visualization only) with waypoints (green) generated from goal (blue). Successes: (a) a valid path under out-of-distribution lunar appearance, (b) threading two rocks with the erosion clearance. Failures: (c) an obstacle forgotten after leaving the field of view, (d) a slope read as a cliff by the flat-ground projection.
}
\label{fig:qual_sim}
\end{figure}

\noindent
\textbf{Is OWLS capable of driving a space rover in a space environment?}
\textit{OWLS reaches the goal on the majority of unseen terrains on both platforms and with both VLMs}: $58.0$ and $60.0$ on Perseverance, $75.0$ and $77.0$ on Pragyan (Table~\ref{table:simulated_space_experiments}).
Fig.~\ref{fig:qual_sim}(a,\,b) shows planning over the VLM feature map handling both open-world appearance and embodiment: on the Moon it plans a valid path over terrain whose appearance lies far outside the VLM's training distribution, and on Mars it threads between two rocks with the clearance the erosion buffer adds.
NoMaD reaches $23.0$ and $35.0$ on the same terrains, behind every OWLS variant and slower on the terrains both solve (Table~\ref{table:simulated_space_experiments}).
OWLS fails by stalling, where the no-progress guard fires, more often than by timing out (36 against 6 of 42 Perseverance failures with GSAM2, 16 against 9 of 25 on Pragyan).
NoMaD's failures also end in the no-progress guard (77 of 77 on Perseverance, 59 of 65 on Pragyan): its trajectory samples are not conditioned on the goal and rarely cover the traversable region toward it, so the selected waypoint makes no progress.

\noindent
\textbf{How much does the choice of VLM matter?}
Under the deployed prompt it makes no measurable difference: the variants land within two points of each other on both platforms ($58.0$ vs.\ $60.0$ and $75.0$ vs.\ $77.0$), despite differing in architecture and training data.
Under other wordings it does: replacing ``\texttt{ground}'' by ``\texttt{area}'' or ``\texttt{region}'' takes the same SAM~3 pipeline to $0.0$ on both platforms (Table~\ref{table:terrain_sweep}), which Sec.~\ref{sec:sensitivity} takes up.

\subsection{Prompt and Morphology Ablations}
\label{sec:sensitivity}

\begin{figure*}[t!]
\centering
\includegraphics[width=0.76\textwidth]{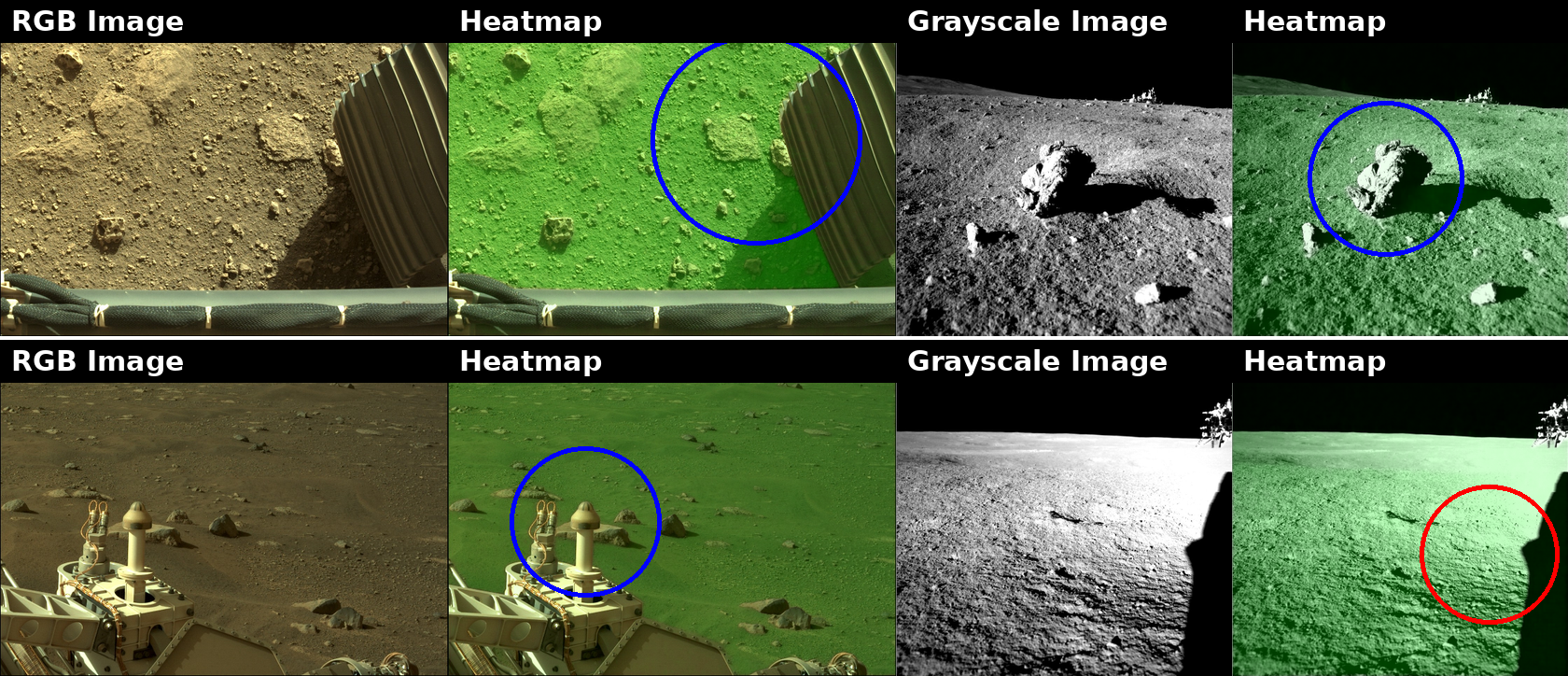}
\caption{
Qualitative results on real planetary rover imagery (Sec.~\ref{sec:offline_inference}). Left: Mars \cite{ai4mars} Right: Moon \cite{isro_navcam}.
Each pair shows the input and the VLM's image-space traversability heatmap (green, brighter is more traversable). Blue circles mark hazards correctly excluded from ground; the red circle marks a shadowed region misclassified as obstacle under extreme lighting.
}
\label{fig:qual_space}
\end{figure*}

The prompt and the two morphological operations are the only inputs an operator sets per deployment, so we vary each over the terrain set of Sec.~\ref{sec:simulated_rover_experiments} with everything else held fixed (Tables~\ref{table:terrain_sweep} and~\ref{table:morphology_scan}).

\noindent
\textbf{Setup.}
The prompt rows replace the single word ``\texttt{ground}'' by five other nouns for free space, or by ``\texttt{rocks}'' to name the hazard instead; Table~\ref{table:morphology_scan} scans each morphological radius under ``\texttt{ground}'' on Perseverance with GSAM2, the other radius held at its deployed value.

\noindent
\textbf{Metric.}
Alongside success rate over the same 100 terrains we log two diagnostics per perception cycle: the maximum VLM score, which is near zero when the VLM cannot ground the prompt, and the fraction of the image the VLM scores above zero, the traversable fraction.
The ``\texttt{ground}'' row repeats Table~\ref{table:simulated_space_experiments} and is the reference for every other row.

\noindent
\textbf{Does the prompt determine the trajectory?}
It does, except on lunar terrain sparse enough that the VLM grounds the hazard noun onto the surface itself.
Because OWLS plans over the VLM's response to the prompt, prompting for ``\texttt{rocks}'' instead of ``\texttt{ground}'' marks hazards as traversable and drops success to $10.0$ (GSAM2) and $34.0$ (SAM~3) on Mars and to $10.0$ for SAM~3 on the Moon.
The exception is GSAM2 on the Moon, $75.0\rightarrow71.0$: with no rock in view, which is most lunar frames, GSAM2 grounds ``\texttt{rocks}'' onto the regolith and its feature map barely changes.

\noindent
\textbf{Does the wording of the prompt matter?}
For GSAM2 it does not; for SAM~3 it decides the outcome.
Replacing ``\texttt{ground}'' by another single free-space noun leaves GSAM2 within four points of its deployed value on Perseverance and eight on Pragyan. The same replacement takes SAM~3 to $0.0$ on both platforms under ``\texttt{area}'' and ``\texttt{region}''.
The diagnostics identify the cause: under those two words SAM~3 returns an empty mask, maximum score zero, on $99$ and $100\,\%$ of Perseverance cycles ($97$ and $93\,\%$ on Pragyan) and proposes no waypoint, whereas GSAM2 scores all six nouns above $0.97$ with a traversable fraction between $0.43$ and $0.64$.
Success falls as the empty-mask rate rises: ``\texttt{path}'' returns an empty mask on $41\,\%$ of Perseverance cycles and $21\,\%$ of Pragyan ones, and scores $15.0$ and $66.0$.
Nor does a higher score identify a better wording: SAM~3 under ``\texttt{surface}'' reaches $90.0$ on Pragyan, above every deployed configuration, with a median peak score of $0.54$ against about $0.1$ for the other nouns, but it does so by calling over $97\,\%$ of the frame traversable, and the same wording leaves Perseverance at $59.0$; we keep ``\texttt{ground}'', fixed before this evaluation.
Language is therefore a working but brittle specification interface: a wording must be validated against the VLM before deployment.

\begin{table}[t]
    \centering
    \setlength{\tabcolsep}{4pt}
    \begin{tabular}{lcccc}
        \toprule
        & \multicolumn{2}{c}{\textit{Perseverance}} & \multicolumn{2}{c}{\textit{Pragyan}} \\
        \cmidrule(lr){2-3} \cmidrule(lr){4-5}
        Condition & GSAM2 & SAM~3 & GSAM2 & SAM~3 \\
        \midrule
        \multicolumn{5}{l}{\textit{Free-space noun}} \\
        \texttt{ground} (deployed) & 58.0 & 60.0 & 75.0 & 77.0 \\
        \texttt{area}          & 58.0 & \phantom{0}0.0 & 74.0 & \phantom{0}0.0 \\
        \texttt{region}        & 54.0 & \phantom{0}0.0 & 83.0 & \phantom{0}0.0 \\
        \texttt{path}          & 60.0 & 15.0 & 73.0 & 66.0 \\
        \texttt{terrain}       & 60.0 & 56.0 & 71.0 & 83.0 \\
        \texttt{surface}       & 59.0 & 59.0 & 73.0 & 90.0 \\
        \quad range over nouns & 54--60 & \phantom{0}0--60 & 71--83 & \phantom{0}0--90 \\
        \multicolumn{5}{l}{\textit{Hazard noun}} \\
        \texttt{rocks}         & 10.0 & 34.0 & 71.0 & 10.0 \\
        \bottomrule
    \end{tabular}
    \caption{
        Prompt ablations (Sec.~\ref{sec:sensitivity}): success rate (\%) over the 100 terrains of Table~\ref{table:simulated_space_experiments}.
    }
    \label{table:terrain_sweep}
\end{table}

\begin{table}[t]
    \centering
    \setlength{\tabcolsep}{6pt}
    \renewcommand{\arraystretch}{0.92}
    \begin{tabular}{lc}
        \toprule
        Radius & Success (\%) \\
        \midrule
        \multicolumn{2}{l}{\textit{Erosion} $\rho_r$} \\
        off                  & 51.0 \\
        $0.50$\,m            & 54.0 \\
        $2.15$\,m (deployed) & \textbf{58.0} \\
        $3.0$\,m             & 52.0 \\
        \multicolumn{2}{l}{\textit{Dilation} $\rho_t$} \\
        off                  & \textbf{59.0} \\
        $0.19$\,m (deployed) & 58.0 \\
        $1.0$\,m             & 45.0 \\
        \bottomrule
    \end{tabular}
    \caption{
        Morphology scan (Sec.~\ref{sec:sensitivity}): GSAM2 success (\%) on Perseverance under \texttt{ground} over the terrains of Table~\ref{table:terrain_sweep}, one radius scanned with the other at its deployed value ($\rho_r=2.15$\,m, $\rho_t=0.19$\,m); bold is best per group.
    }
    \label{table:morphology_scan}
\end{table}

\noindent
\textbf{Are the geometry-set radii effective?}
They are: the radii derived from rover geometry (Sec.~\ref{sec:planning}) match or beat every other value scanned in Table~\ref{table:morphology_scan}, so adapting to a new platform takes geometric measurements rather than parameter tuning.
Erosion success peaks at the deployed $2.15$\,m and falls on either side, by $7$ points with erosion off and $6$ at $3.0$\,m.
Dilation is flat between off and the deployed $0.19$\,m, and a $1.0$\,m radius costs $13$ points, as it erases obstacles the rover cannot clear.
In summary, success falls or plateaus as either radius moves away from its deployed value; as the metric scores no contact, a plateau bounds goal-reaching, not clearance.

\subsection{Ground Detection on Images from Real Space Rovers}
\label{sec:offline_inference}

The VLM feature map is the most failure-sensitive input to the planner, and our modular design lets it be tested on flight imagery without deploying end to end.

\noindent
\textbf{Setup.}
We run SAM~3 \cite{sam3}, without fine-tuning, under the deployed ``\texttt{ground}'' wording on frames from the real Perseverance and Pragyan rovers.
The Mars frames come from Perseverance \cite{ai4mars} and the monochrome Moon frames from the Chandrayaan-3 Pragyan navigation camera \cite{isro_navcam}; none were used to develop or tune the method.

\noindent
\textbf{Metric.}
Labeled flight imagery is too scarce for a quantitative score, so the evidence is visual inspection of the feature maps (Fig.~\ref{fig:qual_space}).

\noindent
\textbf{Is the VLM capable of adapting to sensor noise and extreme lighting from real space rovers?}
Partly: the VLM separates ground from hazards across the domain shift, but misclassifies under extreme lighting (Fig.~\ref{fig:qual_space}); the rocks, robots, and slopes appear as low-scoring regions in the complement of the ground query, not as named concepts.
We hypothesize that extreme lighting is scarce in these models' training data, and a lightweight visibility estimator could mitigate it.

\section{Conclusion}
\label{sec:conclusion}

\noindent
\textbf{Summary.}
We presented OWLS, which plans local navigation directly over the dense feature map of an open-vocabulary VLM: the VLM's per-pixel response to a text prompt is ground-projected into an egocentric value map, adapted to the platform by two morphological operations, and used as the speed field of a minimum-time planner whose trajectory a waypoint controller tracks, with no task-specific training, terrain annotation, or persistent map.
Real-rover trials establish feasibility.
In simulation OWLS reaches the goal on most of 100 unseen terrains per platform with either of two VLMs, ahead of a learned navigation foundation model in all four settings, and naming the hazard instead of free space cuts success in three of four, so the trajectory follows the VLM's language-conditioned output, though rewording the prompt alone can zero it out.
Real planetary imagery shows the VLM transferring across the domain gap while remaining sensitive to extreme lighting.

\noindent
\textbf{Limitations and Future Work.}
We discuss five limitations in our method and evaluation.
First, \textit{the memory-less design} forgets any obstacle leaving the field of view, the most common failure we observe (Fig.~\ref{fig:qual_sim}(c)), which additional cameras or a bounded local memory could mitigate without sacrificing the mapless property.
Second, the method has \textit{no 3D understanding}: the flat-ground assumption behind ground projection (Sec.~\ref{sec:planning}) permits planning from one RGB image without depth sensing, but projects inaccurately on steep slopes (Fig.~\ref{fig:qual_sim}(d)) and likely fails to model overhead clearance and tall obstacles unless augmented with monocular depth or stereo, which we leave for future work.
Third, our method runs an \textit{off-the-shelf VLM} on a compute platform beyond what flight hardware offers today, though a radiation-hardened multicore processor is in development~\cite{hpsc} and a commercial smartphone SoC has already flown on Ingenuity~\cite{ingenuity,snapdragon_jpl}.
Fourth, \textit{language can be a brittle specification interface}: swapping one free-space noun for another can take SAM~3 to zero success (Sec.~\ref{sec:sensitivity}), and current VLMs perform best with prompts representing appearance rather than the physical properties that decide traversal.
Finally, we tested the feature map on flight imagery (Sec.~\ref{sec:offline_inference}) but never closed the loop under \textit{planetary optical effects} in our simulated experiments due to limited modeling of such effects (e.g. the opposition effect washing out texture when the camera faces away from the sun, and the low sun near the lunar poles casting sharp and long shadows).

\iflatexml
\section*{Acknowledgment}
A portion of this research was carried out at the Jet Propulsion Laboratory, California Institute of Technology, under a contract with the National Aeronautics and Space Administration (80NM0018D0004).
\fi

\bibliographystyle{IEEEtran}
\bibliography{references}

\end{document}